\documentclass[a4paper,conference]{IEEEtran}
\ifCLASSINFOpdf
  \usepackage[pdftex]{graphicx}
\else
\fi
\usepackage{xcolor}

\usepackage{amsmath}
\usepackage{bbm}

\usepackage{url}
\usepackage{multirow}
\usepackage[colorinlistoftodos]{todonotes}

\begin{document}
%
% paper title
% Titles are generally capitalized except for words such as a, an, and, as,
% at, but, by, for, in, nor, of, on, or, the, to and up, which are usually
% not capitalized unless they are the first or last word of the title.
% Linebreaks \\ can be used within to get better formatting as desired.
% Do not put math or special symbols in the title.
\title{Deep Convolutional Embedding for Digitized Painting Clustering}

% author names and affiliations
% use a multiple column layout for up to three different
% affiliations
\author{\IEEEauthorblockN{Giovanna Castellano}
\IEEEauthorblockA{Department of Computer Science\\
University of Bari\\
Bari, Italy\\
Email: giovanna.castellano@uniba.it}
\and
\IEEEauthorblockN{Gennaro Vessio}
\IEEEauthorblockA{Department of Computer Science\\
University of Bari\\
Bari, Italy\\
Email: gennaro.vessio@uniba.it}
}

% conference papers do not typically use \thanks and this command
% is locked out in conference mode. If really needed, such as for
% the acknowledgment of grants, issue a \IEEEoverridecommandlockouts
% after \documentclass

% for over three affiliations, or if they all won't fit within the width
% of the page, use this alternative format:
%
%\author{\IEEEauthorblockN{Michael Shell\IEEEauthorrefmark{1},
%Homer Simpson\IEEEauthorrefmark{2},
%James Kirk\IEEEauthorrefmark{3},
%Montgomery Scott\IEEEauthorrefmark{3} and
%Eldon Tyrell\IEEEauthorrefmark{4}}
%\IEEEauthorblockA{\IEEEauthorrefmark{1}School of Electrical and Computer Engineering\\
%Georgia Institute of Technology,
%Atlanta, Georgia 30332--0250\\ Email: see http://www.michaelshell.org/contact.html}
%\IEEEauthorblockA{\IEEEauthorrefmark{2}Twentieth Century Fox, Springfield, USA\\
%Email: homer@thesimpsons.com}
%\IEEEauthorblockA{\IEEEauthorrefmark{3}Starfleet Academy, San Francisco, California 96678-2391\\
%Telephone: (800) 555--1212, Fax: (888) 555--1212}
%\IEEEauthorblockA{\IEEEauthorrefmark{4}Tyrell Inc., 123 Replicant Street, Los Angeles, California 90210--4321}}

% use for special paper notices
%\IEEEspecialpapernotice{(Invited Paper)}

% make the title area
\maketitle

% As a general rule, do not put math, special symbols or citations
% in the abstract
\begin{abstract}
Clustering artworks is difficult for several reasons. On the one hand, recognizing meaningful patterns in accordance with domain knowledge and visual perception is extremely difficult. On the other hand, applying traditional clustering and feature reduction techniques to the highly dimensional pixel space can be ineffective. To address these issues, we propose to use a deep convolutional embedding model for digitized painting clustering, in which the task of mapping the raw input data to an abstract, latent space is jointly optimized with the task of finding a set of cluster centroids in this latent feature space. Quantitative and qualitative experimental results show the effectiveness of the proposed method. The model is also capable of outperforming other state-of-the-art deep clustering approaches to the same problem. The proposed method can be useful for several art-related tasks, in particular visual link retrieval and historical knowledge discovery in painting datasets.
\end{abstract}

% no keywords

% For peer review papers, you can put extra information on the cover
% page as needed:
% \ifCLASSOPTIONpeerreview
% \begin{center} \bfseries EDICS Category: 3-BBND \end{center}
% \fi
%
% For peerreview papers, this IEEEtran command inserts a page break and
% creates the second title. It will be ignored for other modes.
\IEEEpeerreviewmaketitle

\section{Introduction}

Cultural heritage, in particular visual arts, are of inestimable importance for the cultural, historical and economic growth of our societies. In recent years, due to technological improvements and the drastic drop in costs, a large scale digitization effort has been made which has led to an increasing availability of large digitized art collections. Notable examples include WikiArt\footnote{\url{https://www.wikiart.org}} and the MET collection.\footnote{\url{https://www.metmuseum.org/art/collection}} This availability, coupled with the recent advances in Pattern Recognition and Computer Vision, has opened new opportunities to computer science researchers to assist the art community with intelligent tools to analyze and further understand visual arts. Among others, a deeper understanding of visual arts has the potential to make them accessible to a wider population, both in terms of fruition and creation, thus supporting the spread of culture.

The ability to recognize meaningful patterns in visual artworks is intrinsically within the domain of human perception \cite{cupchik2009viewing}. Recognizing the stylistic and semantic attributes of a painting, in fact, arises from the composition of the colour, texture and shape features visually perceived by the human observer. These attributes, which typically relate to color distribution, the spatial complexity of the painted scene, etc., together are responsible for the overall ``visual appearance'' of the artwork \cite{spehr2009image}. Unfortunately, this visual perception can be extremely difficult to conceptualize. However, visual-related features, especially those learned by Convolutional Neural Network (CNN) models \cite{krizhevsky2012imagenet}, can be effective in addressing the problem of automatically extracting useful patterns from the low-level colour and texture features. These patterns can help in various art-related tasks, ranging from object detection in paintings \cite{crowley2014search} to artistic style categorization \cite{van2015toward}. 

Although several successful attempts have been made towards using Pattern Recognition and Computer Vision in art-related supervised tasks (e.g., \cite{cetinic2018fine,garcia2019context,mao2017deepart}), little work has been done in the clustering setting \cite{spehr2009image,barnard2001clustering, gultepe2018predicting}. Having a model that can cluster artworks based on their visual appearance, without the need to collect labels and metadata, can be useful for many applications. It can be used to support art experts in findings trends and influences among painting schools, i.e.~in performing historical knowledge discovery. Analogously, it can be used to discover different periods in the production of the same artist. The model can find out which artworks have mostly influenced the work of current artists. It can support interactive browsing in online art galleries by finding visually linked artworks, i.e.~visual link retrieval. It can help curators better organize permanent or temporary expositions based on their visual similarities rather than historical motivations (for example using indoor navigation systems \cite{piccinni2016distributed}).

In this paper, starting from the deep convolutional embedding clustering (DCEC) model introduced in \cite{guo2017deep}, we propose DCEC-Paint as a method for grouping digitized paintings in an unsupervised fashion. To derive DCEC-Paint we introduced some changes to the original DCEC architecture definition which makes the model more suitable for the specific image domain. We report the results of some experiments, aimed at evaluating the effectiveness of the method in finding meaningful clusters in a dataset of paintings spanning different epochs. The method is also applied to a sub-sample comprising only the works of a single artist, namely Pablo Picasso, to evaluate its effectiveness in finding clusters in the production of a specific artist. Finally, comparative results between the proposed method and other deep clustering approaches to the same problem are reported. 

The rest of this paper is structured as follows. Section 2 deals with related work. Section 3 describes the proposed method. Section 4 and 5 are devoted to the experimental setup and results. Section 6 concludes the work. %and describes some future developments of this research.

\section{Related Work}

In literature, automatic art analysis has been performed using hand-crafted features (e.g., \cite{carneiro2012artistic,khan2014painting,shamir2010impressionism}) or features learned automatically by deep learning models (e.g., \cite{crowley2014search,garcia2019context,mao2017deepart}). Despite the encouraging results of applying feature engineering techniques to this specific domain, early attempts were influenced by the difficulty of capturing explicit knowledge about the attributes to be associated with a particular artist or artwork. This difficulty arises because this knowledge is typically associated with implicit and subjective expertise human experts may find difficult to verbalize. An expert draws his judgment based on the historical context of the work and on understanding the metaphors beyond what is immediately perceived. Furthermore, art experts, as well as untrained enthusiasts, can experience subjective reactions to the stylistic properties of an artwork \cite{cupchik2009viewing}; in other words, emotions can contribute to their aesthetic perception.

In contrast, several successful applications in a number of Computer Vision tasks (e.g., \cite{acharya2019deep,castellano2020crowd,diaz2019dynamically}) have shown that representation learning is an effective alternative to feature engineering for extracting meaningful patterns from complex raw data. In particular, one of the main reasons for the recent success of deep neural network models, such as deep Convolutional Neural Networks, in solving tasks too difficult for classic algorithms is the availability of large human annotated datasets, such as ImageNet \cite{russakovsky2015imagenet}. %The aggregation of all currently available art collections would result in a significantly smaller number of images compared to ImageNet. Instead, 
A model built on these data often provides a sufficiently general knowledge of the ``visual world'' that can be profitably transferred to specific visual domains, in particular the artistic one.

One of the first attempts to use CNNs in the visual art domain was reported in \cite{crowley2014search}. The authors developed a CNN-based system that can learn object classifiers from Google images and use these classifiers to find previously unseen objects in a large painting database. Other works, focusing on object recognition and detection in artworks, were also reported \cite{cai2015cross,cai2015beyond,crowley2016art,gonthier2018weakly,tan2016ceci,wilber2017bam}. The main issue to be faced in this kind of research is the so-called \textit{cross-depiction} problem, that is the problem of recognizing visual objects regardless of whether they are photographed, painted, drawn, etc. The variance between photos and artworks is greater than both domains when considered individually, so classifiers usually trained on traditional photographic images can encounter difficulties when used on painting images, due to the domain shift.

%\todo[inline]{il pezzo che segue lo toglierei, non c'entra molto} In order to reduce the gap between visual features of realistic and artistic data, some works, e.g. \cite{tomei2019art2real,zhu2017unpaired}, proposed image-to-image translation techniques aimed at translating paintings to photo-realistic images. To this end, the Generative Adversarial Network framework is typically used.
%\todo[inline]{}

Another task frequently faced by computer science researchers in this domain is learning to recognize artists by their style. In an early work \cite{van2015toward} van Noord et al.~proposed PigeoNET, a CNN trained on a large collection of paintings to perform the task of automatic artist recognition based on visual characteristics. Classifying the unique characteristics of an artist is a complex task, even for an expert. This is because there can be low inter-variability among different artists and high intra-variability in the style of the same artist. Recently, encouraging results have been reported on the application of deep CNNs to art style classification \cite{cetinic2018fine,mao2017deepart,sandoval2019two}. In other works, e.g.~\cite{strezoski2017omniart,yang2018classifying}, experiments were performed considering also additional metadata, for example time period, reporting better results. 

Another task that has attracted attention is finding similarity relationship between visually linked paintings. In \cite{seguin2016visual}, Seguin et al.~proposed a pre-trained CNN model to predict pairs of paintings that an expert believed had a visual relationship to each other. Similarly, in \cite{shen2019discovering}, Shen et al.~used a deep neural network model to identify near duplicate patterns in a dataset of artworks attributed to Jan Brueghel. 

Most of the existing literature reports the use of machine/deep learning-based solutions that require some form of supervision. Conversely, very little work has been done from an unsupervised perspective. In \cite{barnard2001clustering}, Barnard et al.~proposed a clustering approach to fine art images by exploiting textual descriptions through natural language processing. Spehr et al.~\cite{spehr2009image}, on the other hand, applied a computer vision approach to the problem of grouping paintings using traditional hand-crafted features. Saleh et al.~\cite{saleh2016toward} proposed an unsupervised approach to finding similarities among paintings, based on traditional hand-crafted features. They trained discriminative and generative models for the supervised task of classifying painting style to ascertain what kind of features would be most useful in the artistic domain. Then, once they found the most appropriate features, they used these features to judge the similarity between paintings using distance measures. In \cite{gultepe2018predicting}, Gultepe et al.~applied an unsupervised feature learning method based on $k$-means to extract features which were then fed into a spectral clustering algorithm for the purpose of grouping paintings. In \cite{ircdl} and \cite{MTAP}, we have proposed a method for finding visual links among paintings in a completely unsupervised way. The method relies solely on visual attributes automatically learned by a deep pre-trained model, so it can be particularly effective when additional information, such as textual metadata, are scarce or unavailable. 

The previously described contributions confirm the applicability of a deep learning-based strategy to the problem of visual pattern extraction in painting datasets. Inspired by this success, in this paper we propose to use a deep clustering model to group paintings based on their visual similarity.

\section{Proposed Method}

Clustering is one of the fundamental tasks in Machine Learning. It is notoriously difficult, mainly due to the lack of supervision in evaluating what an algorithm finds. In particular, since its appearance, $k$-means has been widely used due to its ease of implementation and effectiveness \cite{jain2010data}. However, especially in a complex image domain, applying $k$-means may not be feasible. On the one hand, clustering with traditional distance measures in the highly multi-dimensional raw pixel space is well-known to be completely ineffective. Moreover, as noted earlier, extracting meaningful feature vectors based on domain-specific knowledge can be extremely difficult when dealing with artistic data. On the other hand, applying well-known dimensionality reduction techniques, such as PCA \cite{jolliffe2016principal} and matrix factorization \cite{casalino2018framework}, to the original space or to a manually engineered feature space, can ignore possible nonlinear transformations from the original input to the latent space, thus decreasing clustering performance. 

In recent years, a deep clustering paradigm has emerged that exploits the ability of deep neural networks to find complex nonlinear relationships among data for clustering purposes \cite{guo2017deep,xie2016unsupervised,yang2017towards}. The idea is to jointly optimize the task of mapping the input data to a lower dimensional space and the task of finding a set of centroids in this latent feature space.

\begin{figure}[t]
    \centering
    \includegraphics[width=\columnwidth]{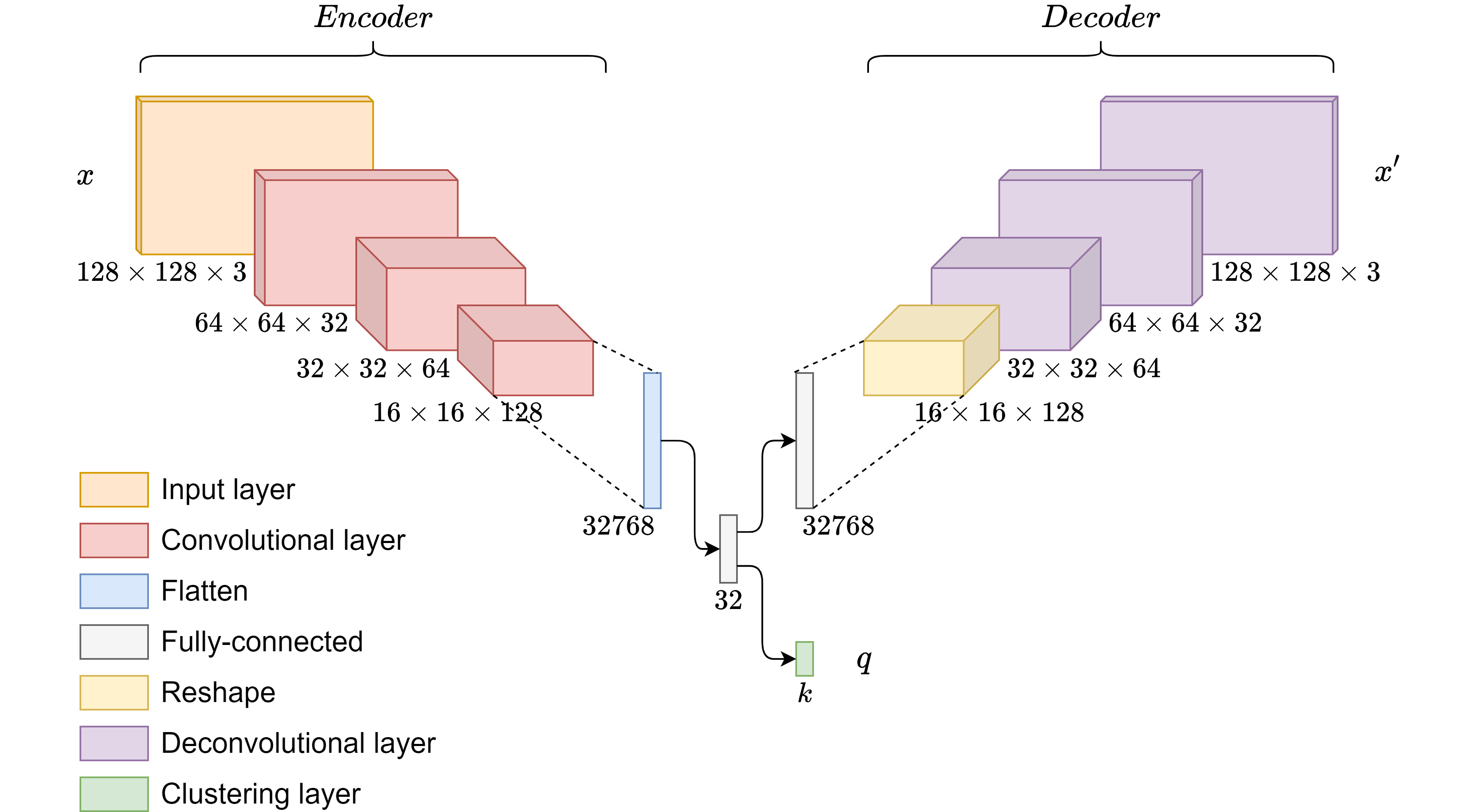}
    \caption{Architecture of DCEC-Paint.}
    \label{fig:model}
\end{figure}

Inspired by the Deep Convolutional Embedding Clustering (DCEC) framework recently proposed by Guo et al.~in \cite{guo2017deep}, we propose DCEC-Paint as a neural network framework for clustering images of digitized paintings. The proposed architecture is depicted in Fig.~\ref{fig:model}. Starting from the DCEC model, we made some architectural changes to the original formulation in order to adapt the model to the specific image domain. In summary: ($i$) the ELU activation function is used instead of ReLU, to speed up learning; ($ii$) the latent embedding space is enlarged to deal with the higher complexity of the input images; ($iii$) the loss importance weights are reversed to give more emphasis to the clustering loss rather than the reconstruction loss.
%The rationale behind these changes is supported by ablation experiments, whose results are reported in the Experimental Section. 

The network is based on a convolutional autoencoder and on a clustering layer attached to the embedded layer of the autoencoder. Autoencoders are neural networks that learn to reconstruct their input \cite{vincent2008extracting,ghosh2019variational}. An autoencoder consists of two parts: an encoder $\phi$, which learns a nonlinear function that maps the input data to a smaller hidden latent space, and a decoder $\psi$, which learns to reconstruct the original input using this latent representation. The autoencoder parameters are updated by minimizing a mean squared reconstruction loss:
\[
\mathcal{L}_r = \frac{1}{n}\sum_{i=1}^{n}\left( x'_i - x_i \right)^2 = \frac{1}{n}\sum_{i=1}^{n}\left( \psi\left( \phi \left( x_i \right) \right) - x_i \right)^2,
\]
where $n$ is the cardinality of the dataset, $x_i$ is the $i$-th input sample and $x'_i$ its reconstruction. We assume an input consisting of $128 \times 128$ three-channel scaled images, normalized in the range $\left[ 0,1 \right]$. This input is then propagated through a stack of convolutional layers that learn to extract hierarchical visual features. The first convolutional layer has $32$ filters, with kernel size $5 \times 5$. The second convolutional layer has 64 filters, with kernel size $5 \times 5$. The third convolutional layer has $128$ filters, with kernel size $3 \times 3$. The number of filters in the last two layers is higher mainly because the number of low level features (i.e., circles, edges, lines, etc.) is typically low, but the number of ways to combine them to get higher level features can be high. All convolutional layers adopt strides $2$ and zero-padding, and are followed by an exponential linear unit (ELU) nonlinearity. We preferred this activation function to the originally proposed ReLU, as ELU tries to make the mean activations closer to zero, thus speeding up learning:
\[
f(x) = 
\begin{cases}
x & \text{ if } x > 0, \\ 
\alpha \left( e^x - 1 \right) & \text{ otherwise, } 
\end{cases}
\]
where $x$ is the input to a neuron and $\alpha=1$ is an extra constant. All units in the last convolutional layer are flattened and given as input to a fully-connected layer with $32$ units, which constitutes the latent embedding space. In the original formulation \cite{guo2017deep}, the number of units in this layer was set to $10$. However, we have found that this dimension is too constraining, making it slower the reconstruction of complex artistic images. The embedding features are then reshaped and propagated through deconvolutional layers, which mirror, in a reverse layer-wise order, the encoder hyper-parameters and restore the embedding features back to the original input.

As in \cite{guo2017deep}, the clustering layer formulation is based on the Deep Embedded Clustering (DEC) proposed in \cite{xie2016unsupervised}. This layer is connected to the bottleneck of the autoencoder and its task is to assign the embedding features of each sample to a cluster. Given an initial estimate of the nonlinear mapping $\phi : X \rightarrow Z$ and initial cluster centroids $\{ \mu_j \}_{j=1}^k$, the clustering layer maps each embedded point, $z_i$, to a cluster centroid, $\mu_j$, using Student's t distribution:
\[
q_{ij} = \frac{\left( 1 + \left \| z_i - \mu_j \right \|^2 \right)^{-1}}{\sum_j \left( 1 + \left \| z_i - \mu_j \right \|^2 \right)^{-1}},
\]
where $q_{ij}$ represents the membership probability of $z_i$ of belonging to cluster $j$; in other words, it can be seen as a soft assignment. The membership probabilities are used to compute an auxiliary target distribution $P$:
\[
p_{ij} = \frac{q_{ij}^2 / \sum_i q_{ij}}{\sum_j \left( q_{ij}^2 / \sum_i q_{ij} \right)},
\]
where $\sum_i q_{ij}$ are soft cluster frequencies. Clustering is performed by minimizing the Kullback-Leibler (KL) divergence between $P$ and $Q$:
\[
\mathcal{L}_c = KL(P \parallel Q) = \sum_i \sum_j p_{ij} \log \left( \frac{p_{ij}}{q_{ij}} \right).
\]
In practice, the $q_{ij}$'s provide a measure of the similarity between each data point and the different $k$ centroids. Higher values for $q_{ij}$ indicate more confidence in assigning a data point to a particular cluster. The auxiliary target distribution is designed to place more emphasis on the data points assigned with higher confidence, while normalizing the loss contribution of each centroid. Then, by minimizing the divergence between the membership probabilities and the target distribution, the network improves the initial estimate by learning from previous high confidence predictions, in a form of self-supervised training.

In \cite{xie2016unsupervised}, the network abandons the decoder and fine-tunes the encoder using only the clustering loss $\mathcal{L}_c$. However, this approach could distort the embedded space, affecting clustering performance. Instead, as in \cite{guo2017deep}, we propose to keep the decoder attached to the encoder during training. This can help DCEC-Paint preserve the data structure of the latent feature space. Overall, the network tries to minimize the following composite objective function:
\[
\mathcal{L} = \lambda \mathcal{L}_r + \left( 1 - \lambda \right)\mathcal{L}_c,  
\]
where $\lambda \in \left[ 0,1 \right]$ is a hyper-parameter that balances $\mathcal{L}_r$ and $\mathcal{L}_c$. In the original formulation \cite{guo2017deep}, $\lambda=0.1$ and the weights were reversed, thus giving more importance to the reconstruction loss than the clustering loss. However, since reconstruction accuracy is not the main focus of the model, we found that placing more emphasis on the clustering term improves cluster assignment. 

The overall training works in two steps. In an initial pre-training phase, the convolutional autoencoder is trained to learn an initial set of embedding features, by miniziming $\mathcal{L}_r$ and keeping $\lambda = 1$. In other words, at this stage, the model behaves only as an autoencoder. After this pre-training, the learned features are used to initialize the cluster centroids $\mu_j$ using the traditional $k$-means. Finally, embedding feature learning and cluster assignment are optimized simultaneously by setting $\lambda=0.1$. Note that optimizing only the clustering loss reduces to the original DEC method. The overall weights are updated using backpropagation. It is worth noting that, to avoid instability, $P$ is not updated on every iteration using only a batch of data, but using all embedded points every $t$ iterations. The training procedure stops when the change in cluster assignments between two consecutive updates is below a given threshold $\delta$. It is worth remarking that ``training'' is understood here as the process of optimizing the reconstruction of the original input, in the case of the autencoder, and the search for cluster centroids, in the case of clustering. Both tasks do not require any form of supervision.

\section{Experimental Setting}

\subsection{Dataset}

To evaluate the effectiveness of the proposed DCEC-Paint method, we used a database that collects paintings of $50$ very popular painters. More precisely, we used data provided by the Kaggle platform,\footnote{\url{https://www.kaggle.com/ikarus777/best-artworks-of-all-time}} scraped from an art challenge website.\footnote{\url{http://artchallenge.ru}} The artists belong to very different epochs and painting schools, ranging from Giotto di Bondone and Renaissance painters such as Leonardo da Vinci and Michelangelo, to more modern exponents, such as Pablo Picasso and Salvador Dal\'i. In particular, nine periods can be recognized: Gothic, Renaissance, Baroque, Romanticism, Impressionism, Post-impressionism, Expressionism, Surrealism, Art Nouveau/Modern Art. Painting images are non-uniformly distributed among painters for a total of $8,446$ images of different sizes. To speed up calculations, each image was resized to $128 \times 128$ pixels; in addition, to improve network performance, the images were normalized in the range $\left[ 0,1 \right]$ before training.

\subsection{Implementation Details}

Experiments were performed on an Intel Core i5 equipped with the NVIDIA GeForce MX110, with dedicated memory of 2GB. As a deep learning framework, we used TensorFlow 2.0 and the Keras API. 

The following section shows the results of some experiments. In the first experiment, we evaluated the effectiveness of DCEC-Paint in clustering the dataset. In addition, we run our method on a sub-sample of paintings belonging to Pablo Picasso. This was done to evaluate the effectiveness of the proposed method in the search for meaningful clusters within the production of the same artist. In the third experiment, we compared the proposed solution with its original formulation to justify the change we made to the loss weights. Finally, we fairly compared the proposed method with other deep clustering approaches, to assess whether it provides a better solution to the problem of clustering paintings. In particular, we considered the following two alternative approaches: 
\begin{enumerate}
\item
Running $k$-means on the embedded features of the proposed pre-trained convolutional autoencoder (CAE), hereafter referred to as CAE+$k$-means; 
\item
The Deep Embedding Clustering (DEC) method proposed by Xie et al.~\cite{xie2016unsupervised}, in which, after the pre-training stage, the decoder is abandoned and only the clustering loss is minimized. It is worth noting that, for a fair comparison, DEC was not set up as a fully-connected multi-layer perceptron as in \cite{xie2016unsupervised}, but mirrored the same architecture as the proposed CAE. 
\end{enumerate}
%visto fin qui

In all cases, CAE has been pre-trained end-to-end for $200$ epochs using the AdaMax optimizer and mini-batches of size $128$. To initialize the cluster centroids, we run $k$-means with 20 restarts, choosing the best solution. For DEC and DCEC-Paint, the convergence threshold $\delta$ has been set to $0.001$ and the update interval $t$ to $140$. 

\subsection{Evaluation Metrics}

Since clustering is unsupervised, we do not know \textit{a priori} which is the best grouping of paintings. Furthermore, since two artworks by the same artist could have been produced in different stylistic periods, it is very difficult to assign a precise label to a given painting, thus providing a form of supervision over cluster assignments. For this reason, for clustering evaluation, we mainly used two standard internal metrics, i.e.~the silhouette coefficient \cite{rousseeuw1987silhouettes} and the Calinski-Harabasz index \cite{calinski1974dendrite}, which are based on the model itself. The silhouette coefficient is defined for each sample and is calculated as follows:
\[
SC = \frac{b-a}{max(a,b)},
\]
where $a$ is the mean distance between a data point and all other points in the same cluster, and $b$ is the mean distance between a data point and all other points in the nearest cluster. The final score is obtained by averaging over all data points. The silhouette coefficient is bounded between $-1$ and $1$, which represent the worst and best possible value, respectively. Values close to $0$ indicate overlapping clusters. 

The Calinski-Harabasz index is the ratio of the sum of between-cluster dispersion and inter-cluster dispersion for all clusters. More precisely, for a dataset $D$ of size $n_D$, which has been partitioned into $k$ clusters, the index is defined as:
\[
CHI = \frac{\mathrm{tr}(B_k)}{\mathrm{tr}(W_k)} \times \frac{n_D - k}{k - 1},
\]
where $\mathrm{tr}(B_k)$ is the trace of the between group dispersion matrix and $\mathrm{tr}(W_k)$ is the trace of the within-cluster dispersion matrix. These matrices are defined as follows:
\[
W_k = \sum_{q=1}^k \sum_{x \in C_q} (x - c_q) (x - c_q)^T,
\]
\[
B_k = \sum_{q=1}^k n_q (c_q - c_D) (c_q - c_D)^T,
\]
where $C_q$ is the set of points in cluster $q$, $c_q$ the center of cluster $q$, $c_D$ the center of $D$, and $n_q$ the cardinality of cluster $q$. It is worth noting that the Calinski-Harabasz index is not bounded within a given interval, but its value tends to grow. For this reason, only relative values normalized by the maximum value obtained are shown below.

It is worth remarking that both $SC$ and $CHI$ were computed in the space induced by the embedding. Estimating these metrics in the original high dimensional space would have been extremely problematic.

The above metrics are based on internal criteria. However, note that the nine periods into which the dataset we used can be divided provide a form of ground truth. Moreover, paintings can be further divided into two macro periods if we split them between more classic Pre-Impressionism works and more modern (Post-)Impressionism paintings. In this way, we can also calculate the unsupervised clustering accuracy, which is widely used in an unsupervised setting:
\[
ACC = max_n \frac{\sum_{i=1}^{n} 1 \{l_i = m(c_i)\}}{n},
\]
where $l_i$ is the ground-truth label, $c_i$ is the cluster assignment, and $m$ varies over all possible one-to-one mappings between clusters and labels.

Finally, we also drew qualitative observations on the cluster assignments provided by the method.
%Finally, we evaluate the cluster assigniments of the model  

\section{Results}

The results obtained from the various experiments are reported in the following subsections.
%by clustering the overall dataset and those obtained by focusing only on the artworks produced by a single artist, namely Pablo Picasso. Moreover, we report the results of an ablation study and those obtained by comparing the proposed method with other state-of-the-art approaches.

\subsection{Overall Dataset}

\begin{figure}[t]
    \centering
    \includegraphics[width=\columnwidth]{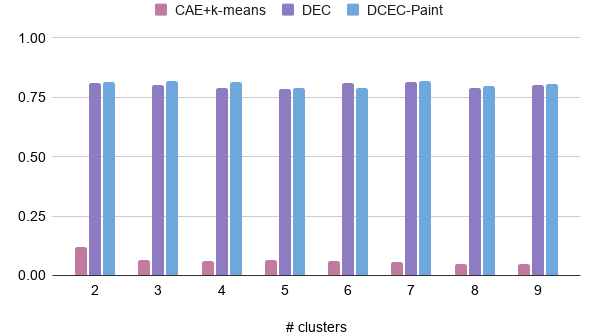}
    \caption{Silhouette coefficient on the overall dataset.}
    \label{fig:sc}
\end{figure}

\begin{figure}[!t]
    \centering
    \includegraphics[width=\columnwidth]{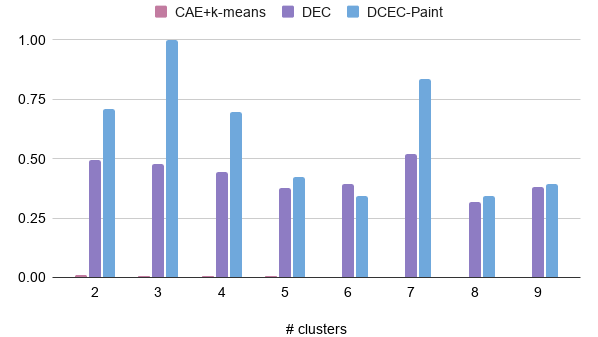}
    \caption{Calinski-Harabasz index on the overall dataset.}
    \label{fig:chi}
\end{figure}

\begin{figure}[t]
    \centering
    \includegraphics[width=\columnwidth]{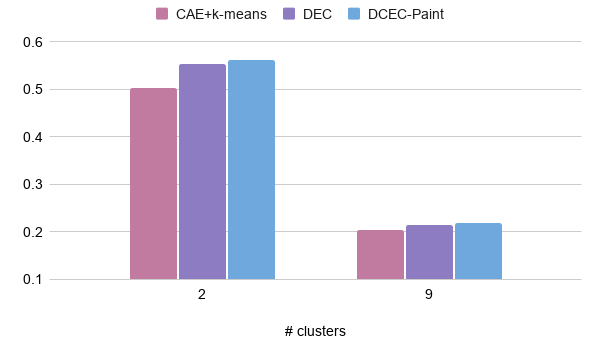}
    \caption{Unsupervised  clustering  accuracy on the overall dataset.}
    \label{fig:acc}
\end{figure}

Figures \ref{fig:sc}--\ref{fig:acc} show the clustering performance of the proposed DCEC-Paint on the entire dataset by varying the number of clusters $k$. Note that, for reasons of space, the figures also show the results obtained with other methods, which will be discussed in a subsequent subsection. We varied $k$ between $2$, which is the minimum number of clusters, and $9$, which is the grouping suggested by the nine different painting schools to which the artworks in the dataset historically belong. By observing the silhouette coefficient, it can be seen that well-defined clusters are obtained in all cases, with the two highest values at $k=3$ and $k=7$. The values of the Calinski-Harabasz index tend to increase or decrease accordingly. Regarding unsupervised clustering accuracy, the method achieved an accuracy of $\sim0.56$ in the binary discrimination and of $\sim0.22$ in the case of nine clusters. Time period classification is notoriously difficult (see, for example, \cite{chen2019recognizing} and \cite{saleh2015large}), as more classic works may exhibit futuristic and pioneering features, while modern works can draw inspiration and revive the classic style. Furthermore, these rather low results suggest that the model tends to look at content rather than stylistic features to group paintings.

\begin{figure}[t]
    \centering
    \includegraphics[width=\columnwidth]{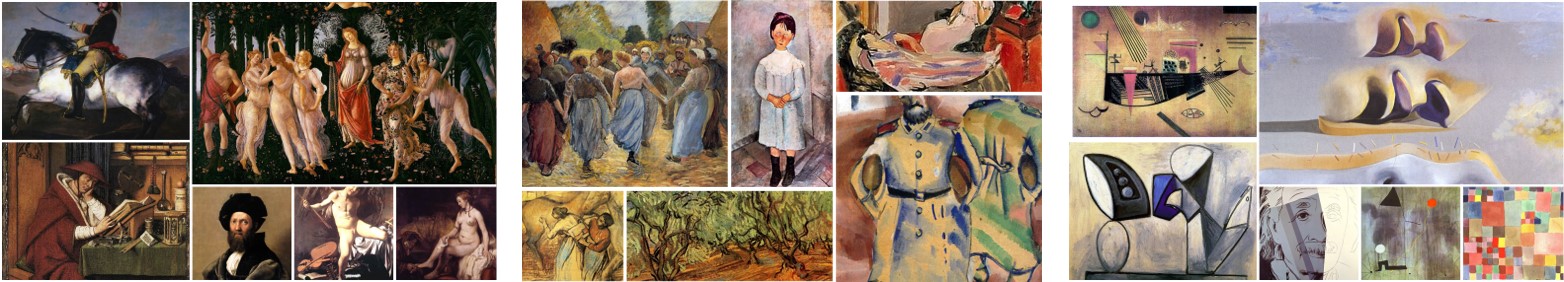}
    \caption{Sample images from clusters when $k=3$.}
    \label{fig:dcec_3}
\end{figure}

\begin{figure}[t]
    \centering
    \includegraphics[width=\columnwidth]{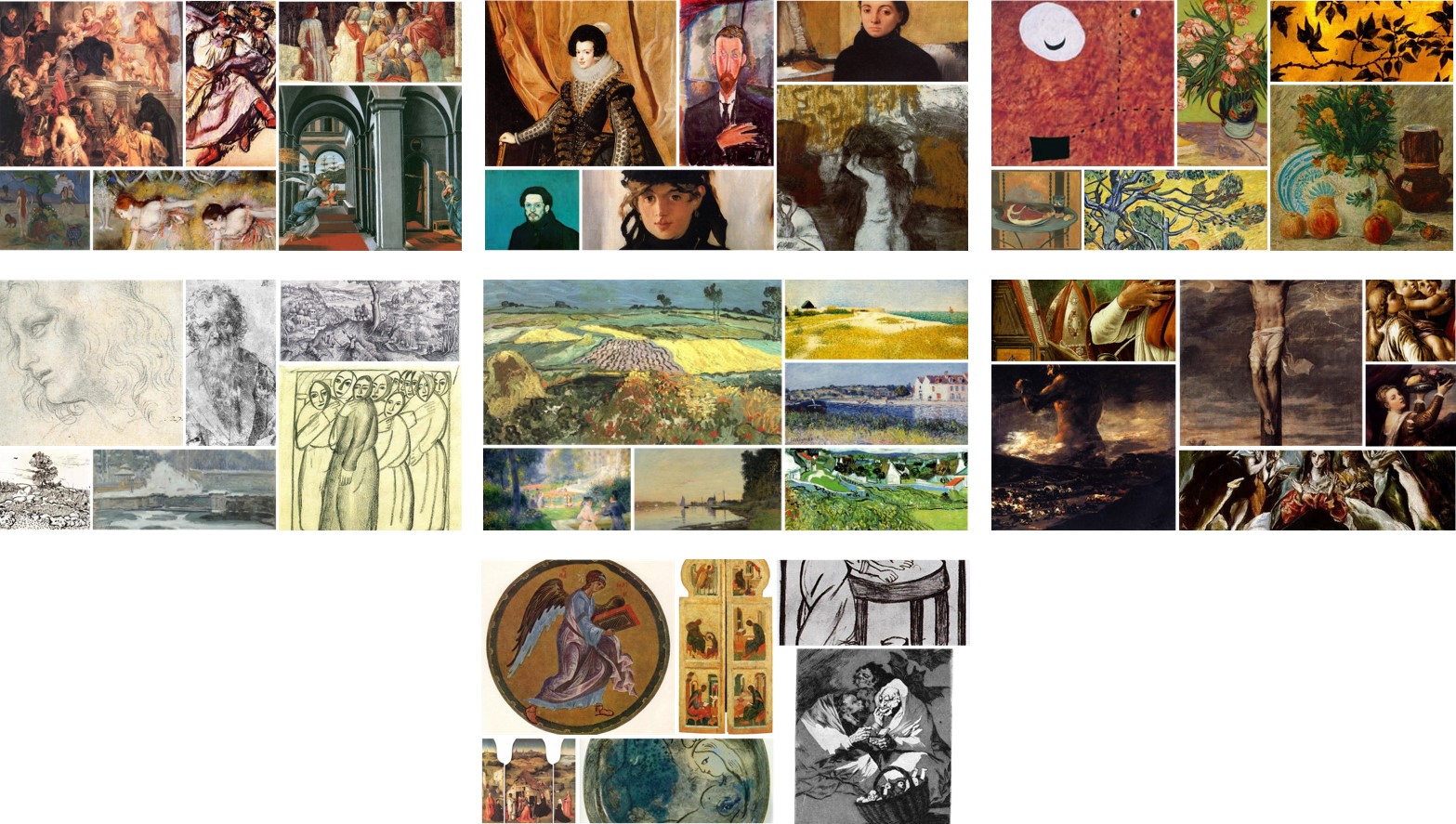}
    \caption{Sample images from clusters when $k=7$.}
    \label{fig:dcec_7}
\end{figure}

From a qualitative point of view, Fig.~\ref{fig:dcec_3} and \ref{fig:dcec_7} show sample images from the clusters obtained with DCEC-Paint when $k=3$ and $k=7$. In the case of three clusters, the cluster assignment suggests that the model was able to some extent to separate the artworks into three macro-periods: (\textit{i}) more classic works, including Renaissance, Romanticist and Baroque paintings; (\textit{ii}) artworks mostly belonging to the Impressionist and Post-Impressionist period, such as paintings of van Gogh and Degas; (\textit{iii}) more modern samples, including works by Picasso and Dal\'i. In other words, with this low number of clusters, the model mainly looked at the stylistic attributes of paintings to group them. Conversely, by increasing the number of clusters to $7$, it is more likely to find works from very different periods in the same clusters but sharing some other visual characteristics. In fact, in the case of $7$ clusters, the model appears to have examined seven distinctive features. Two clusters appear to be related to people: groups of several individuals in one cluster; and single individuals, typically in portraits, in the other cluster. Another cluster mainly contains drawings: the dataset, in fact, includes several drawings by Da Vinci, Duerer, and so on. One cluster mainly contains landscapes, regardless of the stylistic school. Another cluster is made up of iconic works, mostly from the Gothic period. One cluster seems to be related to dark, Romanticist scenes. Finally, one cluster seems to include still-life paintings, flowers and, more generally, household items. These findings suggest that as the number of clusters increases, the model begins to use content-based features to group artworks, thus confirming the results obtained with unsupervised accuracy.

After training the two best models until convergence, we used the learned embedding features to fit t-distributed Stochastic Neighbor Embedding (t-SNE) representations for the purpose of data visualization. t-SNE is a nonlinear dimensionality reduction technique suitable for embedding high-dimensional data in a low-dimensional space of two or three dimensions \cite{maaten2008visualizing}. The graphical representation given in Fig.~\ref{fig:t-sne} confirms the effectiveness of the model in finding clusters that appear to be well-separated.

%\setlength{\tabcolsep}{4pt}
%\begin{table}[t]
%\centering
%\label{tab:clustering}
%\caption{Performance of DCEC-Paint on the Overall Dataset.}
%\begin{tabular}{lll}
%\hline
%$\mathbf{k}$ & $\mathbf{ss}$ & $\mathbf{chs}$ \\
%\hline
%\multirow{2}{*}{\textbf{$k$}} & \multicolumn{2}{l}{\textbf{Proposed DCEC}}   \\
                           %           & \textit{\textbf{ss}} & \textit{\textbf{chs}} \\ \hline
%2                                     & 0.8164               & 0.7092                \\
%3                                     & 0.8207               & 1.0000                 \\
%4                                     & 0.8155               & 0.6965                \\
%5                                     & 0.7911               & 0.4235                \\
%6                                     & 0.7905               & 0.3434                \\
%7                                     & 0.8186               & 0.8340                \\
%8                                     & 0.7988               & 0.3415                \\
%9                                     & 0.8058               & 0.3916                \\ \hline
%\end{tabular}
%\end{table}
%\setlength{\tabc%olsep}{1.4pt}

\subsection{Single Artist Data}

\begin{figure}[t]
    \centering
    \includegraphics[width=\columnwidth]{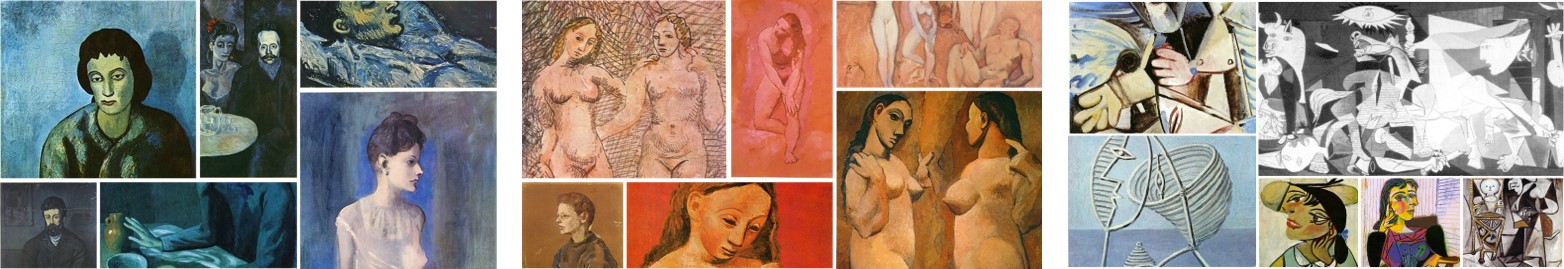}
    \caption{Sample images from Picasso clusters.}
    \label{fig:picasso_3}
\end{figure}

\begin{figure}[t]
    \centering
    \includegraphics[width=.3\columnwidth]{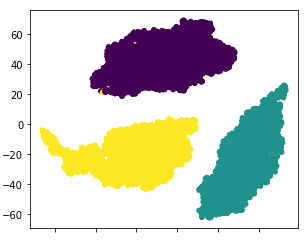}
    \includegraphics[width=.3\columnwidth]{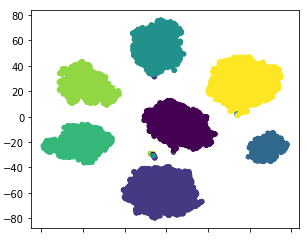}
    \includegraphics[width=.3\columnwidth]{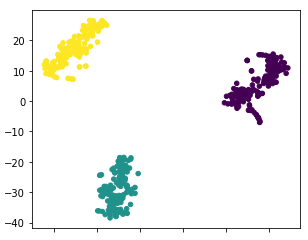}
    \caption{From left to right, t-SNE visualizations of the clusters in Fig.~6--8.}
    \label{fig:t-sne}
\end{figure}

We also run DCEC-Paint on the $439$ artworks painted by Pablo Picasso the dataset we used was provided with. We set $k=3$ because historically three clearly distinguishable macro-periods can be recognized in Picasso's artistic production: blue period; rose period; and cubism. Although forms of proto-cubism can be traced in the first two periods, marking a transition from earlier works towards more mature production, the three periods present evident stylistic (and color) differences. 
%In performing the clustering, DCEC-Paint achieves a high value of $ss$ equals to $0.9356$ and a $chs$ of $10730.87$. The latter value is not normalized as it concerns a much smaller dataset. These performance, together with the corresponding t-SNE visualization of the embedding features learned by the model (Fig. \ref{fig:picasso_3}), shows that the proposed approach is really effective in finding well-defined clusters. 
Figure \ref{fig:picasso_3} shows sample images from the clusters obtained with the method. Thanks to the very different color distribution, the model was quite good at grouping works belonging to the same stylistic period of the artist. Figure \ref{fig:t-sne} confirms that the proposed approach is really effective in finding well-defined clusters.

\begin{figure}[!t]
    \centering
    \includegraphics[width=\columnwidth]{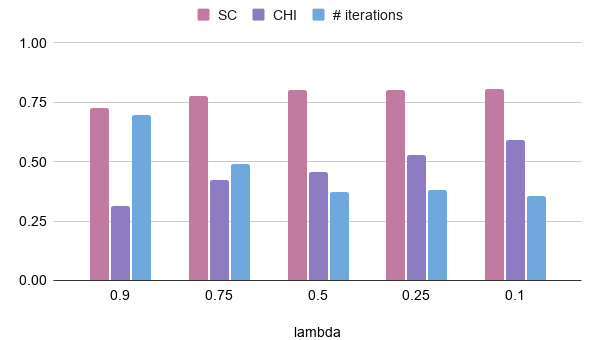}
    \caption{Results of the ablation study. The values are averaged over the different values for $k$. The number of iterations is expressed in tens of thousands.}
    \label{fig:ablation}
\end{figure}

\subsection{Ablation Study}

%In this subsection, we report the results of the comparison between the original formulation of DCEC, provided with an embedded space of $10$ features, and our proposal, in which the embedding features are augmented to $32$. Figure \ref{fig:reconstruction} plots the reconstruction loss against the $200$ epochs of the pre-training stage. We limited our analysis only to this stage, as the reconstruction of the input images was not the primary goal of the study, so we were not interested in the behaviour of the model in the following epochs. As it can be observed, the proposed DCEC takes on more erratic values at earlier epochs, but exhibits slightly better convergence properties at later epochs. This can be explained considering that a slightly larger embedded space may have more representation power useful for the reconstruction of the complex art images.
%In addition, 

We studied the effects of the loss weights assigned to the composite loss function $\mathcal{L}$ on clustering performance. In the original formulation of DCEC, the weights are reversed and the joint loss takes the following form: $\mathcal{L}=(1-\lambda)\mathcal{L}_r + \lambda \mathcal{L}_c$, with $\lambda$ evaluating $0.1$. This form puts more emphasis on the reconstruction loss than the clustering loss during backpropagation. However, since the accurate image reconstruction is not the primary task of the model, we reversed the weights. Figure \ref{fig:ablation} shows the results in terms of $SC$, $CHI$ and number of iterations before reaching convergence, averaged over the different values for the number of clusters $k$ (from 2 to 9), obtained by varying the value of the weight parameter $\lambda$. Note that decreasing the value for $\lambda$ progressively gives more weight to the clustering loss rather than to the reconstruction loss. 
%between the original DCEC and the proposed one, by varying the number of clusters $k$. Note that also the results of applying no weights, i.e. using a joint loss in which the reconstruction and clustering term have the same importance, are reported. 
As can be seen, there is a trend in which gradually giving more importance to the clustering term rather than the reconstruction term improves prediction performance, while reducing computational cost.

\subsection{Comparison with SoA}

Figures \ref{fig:sc}--\ref{fig:acc} show the comparison between DCEC-Paint and CAE+$k$-means and DEC. The clustering performance of CAE+$k$-means clearly indicates that this approach is completely ineffective, with performance decreasing as $k$ increases. DEC compares favorably with our approach with fairly similar values for $SC$, and lower performance in terms of $CHI$. Both DCEC-Paint and DEC agree that the partitions into $3$ and $7$ clusters are among the best solutions, even though the second highest performance of DEC is at $k=2$. A different behavior can be observed in the $ACC$ values obtained by CAE+$k$-means. Although, in accordance with internal evaluation criteria, this simple method is not able to find well-defined clusters, it is nevertheless able to exploit some homogeneous properties related to the time period.

Note that, since $k$-means only finds convex clusters, we also used other more sophisticated clustering methods, such as spectral clustering \cite{von2007tutorial}. However, we found that they do not provide significant improvements on the same data; moreover, they are much more computationally demanding.

\section{Conclusion}

The contribution of this study was the achievement of new results in the automatic analysis of artworks, which is a very difficult task. In fact, recognizing meaningful patterns in paintings in accordance with domain knowledge and human visual perception is extremely difficult for machines. For this reason, applying traditional clustering and feature reduction techniques to the highly dimensional pixel space has been largely ineffective. To address these issues, we proposed to use a deep convolutional embedding clustering model that relies only on visual features automatically learned by the deep network model. The model was able to find well-separated clusters both when considering an overall dataset spanning different epochs and when focusing on works produced by the same artist. Quantitative and qualitative results confirmed the effectiveness of the method. In particular, from a qualitative point of view, it seems that the model is able to recognize stylistic or semantic attributes of paintings to group them. When the granularity of clustering is coarse, the model takes into account more general features, mainly related to the artistic style. When the granularity is finer, the model begins to use content features and tends to group works regardless of the corresponding painting school. 

Although this model is not completely new, we succeeded in finding a suitable variant of the original model so that, once appropriate features are automatically extracted from the images of paintings, it can help to acquire new knowledge about the relationships among paintings, useful for several applications, in a completely unsupervised way. This kind of knowledge extraction has been neglected in previous studies on artwork clustering, where traditional approaches have been used. The results obtained surpass previous studies, as we took into account the ability of convolutional neural network models to exploit complex nonlinear relationships within data. The results obtained are encouraging for the purposes of our research, whose long-term goal is the automatic discovery of patterns in painting images without the need of prior knowledge and labels and metadata, which are very difficult to collect in this domain, even for an expert. In future work, we wish to integrate (con)textual information---users can provide simply by looking at an artwork without prior knowledge, in order to try to mimic the complex human aesthetic perception. Finally, we also want to discard traditional distance measures to find clusters in the feature space, relying on a metric learning approach.

\section*{Acknowledgment}

Gennaro Vessio acknowledges funding support from the Italian Ministry of Education, University and Research through the PON AIM 1852414 project.

%The authors would like to thank...

% trigger a \newpage just before the given reference
% number - used to balance the columns on the last page
% adjust value as needed - may need to be readjusted if
% the document is modified later
%\IEEEtriggeratref{8}
% The "triggered" command can be changed if desired:
%\IEEEtriggercmd{\enlargethispage{-5in}}

% references section

% can use a bibliography generated by BibTeX as a .bbl file
% BibTeX documentation can be easily obtained at:
% http://mirror.ctan.org/biblio/bibtex/contrib/doc/
% The IEEEtran BibTeX style support page is at:
% http://www.michaelshell.org/tex/ieeetran/bibtex/
\bibliographystyle{IEEEtran}
% argument is your BibTeX string definitions and bibliography database(s)
\bibliography{IEEEabrv,biblio}
%
% <OR> manually copy in the resultant .bbl file
% set second argument of \begin to the number of references
% (used to reserve space for the reference number labels box)
%\begin{thebibliography}{1}

%\bibitem{IEEEhowto:kopka}
%H.~Kopka and P.~W. Daly, \emph{A Guide to \LaTeX}, 3rd~ed.\hskip 1em plus
%  0.5em minus 0.4em\relax Harlow, England: Addison-Wesley, 1999.

%\end{thebibliography}

% that's all folks
\end{document}